%% file: main.tex
\documentclass[10pt,twocolumn,letterpaper]{article}
\usepackage[letterpaper,left=.68in,right=.68in,top=.77in,bottom=.83in,columnsep=.29in]{geometry}
\usepackage{mathptmx}
\usepackage[T1]{fontenc}
\usepackage[utf8]{inputenc}
\usepackage{microtype}
\usepackage{amsmath,amssymb}
\usepackage{graphicx}
\usepackage[hyphens]{url}
\usepackage{booktabs,array,tabularx,multirow}
\usepackage{algorithm}
\usepackage[noend]{algpseudocode}
\usepackage{caption}
\usepackage{float,placeins}
\usepackage[numbers,sort&compress]{natbib}
\usepackage[hidelinks]{hyperref}
\date{}
\hypersetup{
  pdftitle={AdvMTO: Multimodal and Temporal Adversarial Object Attacks via 2D Gaussian Splatting for BEV-Based 3D Object Detection},
  pdfauthor={Zhiyang Cui, Hao Tan, Ruonan Li, Shibin Wu, Junjian Zhang, Hong'an Lin, Zhaoquan Gu}
}
\title{AdvMTO: Multimodal and Temporal Adversarial Object Attacks via 2D Gaussian Splatting for BEV-Based 3D Object Detection}
\author{%
  \textbf{Zhiyang Cui}\textsuperscript{1}\quad \textbf{Hao Tan}\textsuperscript{1}\quad
  \textbf{Ruonan Li}\textsuperscript{2}\quad \textbf{Shibin Wu}\textsuperscript{1}\\[3pt]
  \textbf{Junjian Zhang}\textsuperscript{3}\quad \textbf{Hong'an Lin}\textsuperscript{1}\quad
  \textbf{Zhaoquan Gu}\textsuperscript{2,1}\\[6pt]
  \normalfont\normalsize\textsuperscript{1}Harbin Institute of Technology\qquad
  \textsuperscript{2}Peng Cheng Laboratory\\
  \normalfont\normalsize\textsuperscript{3}National University of Defense Technology
}

\begin{document}

\maketitle

\begin{abstract}
Adversarial object attacks pose a threat to BEV-based 3D object detectors
in autonomous driving. Existing attacks typically target a single modality,
coordinate camera and LiDAR observations at the current timestamp, or model
temporal persistence mainly in camera streams. Without a unified
construction of current and historical observations, attacks can leave
part of the detector's accumulated evidence unsuppressed, limiting their
effectiveness. In this paper, we propose AdvMTO, a 2D Gaussian Splatting
(2DGS) framework for suppressing detection of persistent adversarial objects.
Its oriented surfels encode appearance, geometry, and opacity in one object
state, from which camera images and scanline LiDAR returns are generated.
AdvMTO then constructs current and historical inputs through detector-native
processing, regenerating LiDAR returns in each native source sweep for
fusion detectors. We optimize bounded appearance perturbations for each
detector, propagating gradients through historical BEV memory for temporal
camera detectors. For fusion detectors, bounded geometry refinement along
local normals further perturbs both sensing modalities. Experiments on
nuScenes demonstrate effective suppression of traffic-cone detection across
three BEV detectors. Temporal ablations further highlight the importance
of detector-native history modeling for both attack construction and
optimization.
\end{abstract}

\section{Introduction}

Bird's-eye-view (BEV) representations provide a common spatial framework
for 3D object detection in autonomous driving, supporting perception from
camera images, LiDAR point clouds, or both
\cite{philion2020lift,yin2021centerpoint,liu2023bevfusion,li2022bevformer}.
Although sensor fusion and temporal aggregation improve perception,
they do not ensure robustness to adversarial objects
\cite{zhu2023bevrobustness,cheng2024fusion}. A persistent adversarial
object can affect both current observations and the historical evidence
accumulated as the vehicle moves. Evaluating such attacks therefore
requires considering how the same object is observed across sensing
modalities and over time.

Existing object-level attacks span single-modal, multimodal, and temporal
settings. Single-modal attacks modify surface appearance to deceive
camera-based perception \cite{wang2022fca,suryanto2022dta} or alter
object geometry to suppress LiDAR-based detection
\cite{tu2020physically}.
Multimodal attacks construct adversarial 3D objects that jointly
affect camera and LiDAR perception
\cite{abdelfattah2021adversarial,tu2022multisensor}, with MSF-ADV
coupling the two modalities through a deformable mesh
\cite{cao2021lidaradvmulti}.
Temporal attacks seek consistent adversarial effects across views
and consecutive frames \cite{zhu2023bevrobustness}.
SABER, for example, maintains a scene-fixed adversarial object
across camera frames to disrupt the detection of other vehicles
\cite{li2026saber}.
Existing multimodal object attacks typically coordinate camera
and LiDAR observations at the current timestamp, while temporal
object attacks mainly model persistence in camera streams.
We therefore aim to suppress detection of persistent adversarial
objects across sensing modalities and over time.

For a persistent adversarial object, multimodal consistency must
extend from the current frame to the detector's historical inputs.
LiDAR returns depend on the acquisition geometry of each source sweep,
while historical camera observations influence the current prediction
through BEV memory.
Historical observations can retain evidence of the object even when its
current observation is perturbed, limiting attack effectiveness.
The challenge is to construct consistent current and historical observations
across sensing modalities and account for their accumulated influence during
optimization. This requires an editable 3D object state from which both
camera images and LiDAR returns can be generated.
Standard 3D Gaussian Splatting (3DGS) uses volumetric primitives whose
orientations do not directly define surface normals, whereas 2DGS uses
oriented planar primitives with
explicit local surface directions \cite{kerbl20233dgs,huang20242dgs}.
These planar primitives support camera rendering and scanline LiDAR
return generation, while their local normals guide bounded geometry
refinement.

To this end, we propose AdvMTO, a 2DGS-based framework
for suppressing detection of persistent adversarial objects.
Its oriented surfels encode appearance, geometry, and opacity in one
object state.
Within each sequence, the object remains fixed in world coordinates
while its observations follow the recorded sensor poses.
For fusion detectors, AdvMTO regenerates object returns in every native
LiDAR source sweep before temporal alignment, while temporal camera
detectors process current and historical camera observations through
their native temporal pathway.
We optimize bounded appearance perturbations separately for each detector.
For temporal camera detectors, gradients propagate through BEV-memory
updates, allowing the current attack objective to guide appearance
optimization across historical renderings.
Bounded geometry refinement along local normals further perturbs both
sensing modalities for fusion detectors.

Our main contributions are:
\begin{itemize}
    \item We introduce AdvMTO, a 2DGS framework that converts one editable
    object state into camera images and LiDAR returns, with local surface
    normals guiding bounded geometry refinement.

    \item We develop detector-native temporal construction and optimization.
    LiDAR returns are regenerated at each source pose for fusion detectors,
    while gradients propagate through BEV memory and historical renderings
    for temporal camera detectors.

    \item Temporal ablations show that LiDAR history construction changes
    both clean detectability and measured attack effectiveness, while
    backpropagation through BEV memory improves suppression over
    stop-gradient optimization under the same adversarial histories.
    Across three BEV detectors on nuScenes, 2DGS attacks also achieve
    larger relative reductions in traffic-cone target scores than 3DGS attacks.
\end{itemize}

\begin{figure*}[!t]
\centering
\includegraphics[width=\textwidth]{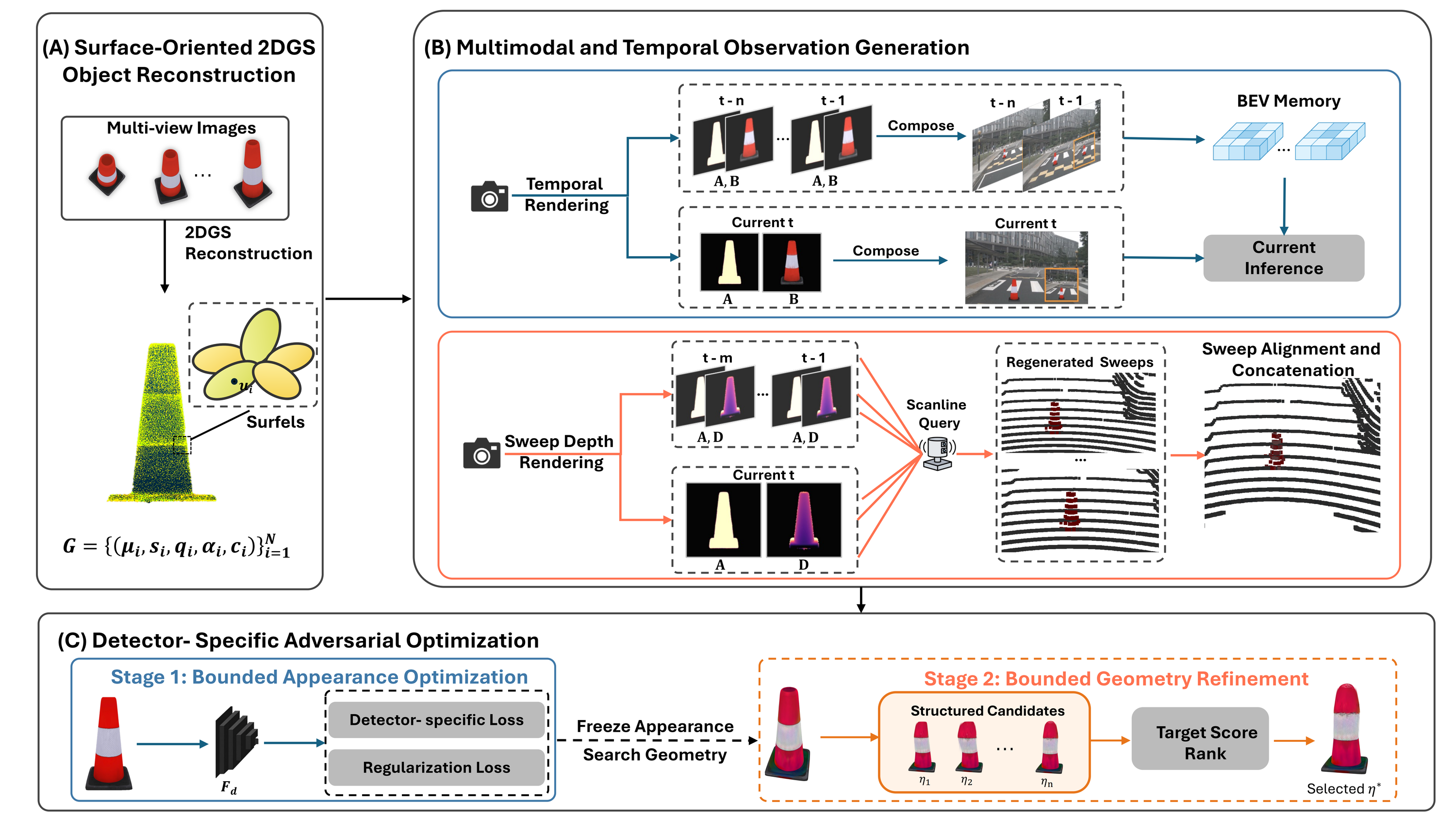}
\caption{Overview of AdvMTO: (A) surface-oriented 2DGS representation,
(B) camera and native-sweep LiDAR generation with detector-native temporal
processing, and (C) bounded appearance optimization and geometry refinement.}
\label{fig:framework}
\end{figure*}

\section{Related Work}

\paragraph{Adversarial Attacks on Autonomous-Driving Perception.}
Adversarial attacks on autonomous-driving perception differ in the
sensor inputs and temporal relationships they manipulate.
Camera-based attacks optimize surface camouflage under varying
viewpoints and environmental conditions
\cite{zhang2019camou,wang2022fca,suryanto2022dta}.
Further work considers visual naturalness and transferability
\cite{wang2021das,suryanto2023active}, and addresses conflicting
texture gradients across viewpoints \cite{liang2025grac}.
Related physical attacks also target monocular and stereo depth
estimation
\cite{yamanaka2020depthpatch,cheng2022physicaldepth,zhao2026cheatingstereo}.
LiDAR attacks manipulate object geometry through adversarial
3D objects \cite{tu2020physically}, or alter sensor returns
through point injection and physical removal
\cite{cao2019lidarattack,cao2023physicalremoval}.

Multi-sensor attacks, including MSF-ADV, construct adversarial
objects that jointly affect camera and LiDAR perception
\cite{cao2021lidaradvmulti,abdelfattah2021adversarial,tu2022multisensor},
with further work targeting camera--LiDAR--radar fusion
\cite{zhu2024malicious}.
Camera-only patches can also compromise camera--LiDAR fusion
models \cite{cheng2024fusion}.
For BEV perception, Zhu et al.\ introduce spatially consistent
and temporally universal patch attacks \cite{zhu2023bevrobustness}.
SABER places a scene-fixed adversarial object to disrupt the
detection of other vehicles, accounting for perspective and
occlusion across views and frames \cite{li2026saber}.
Temporal-misalignment attacks instead manipulate synchronization
between sensor streams \cite{shahriar2026temporal}.
AdvMTO aims to suppress detection of the adversarial object itself.
It generates camera and LiDAR observations from the same object state
and incorporates historical observations through each detector's native
input pathway.

\paragraph{Object Representations and Multimodal Sensor Rendering.}
Object representations determine how appearance and geometry can be
optimized and converted into sensor observations. MeshAdv uses
differentiable rendering to optimize adversarial mesh geometry and texture
\cite{xiao2019meshadv}, while mesh-based multi-sensor attacks couple camera
and LiDAR effects through an explicit object surface
\cite{cao2021lidaradvmulti,abdelfattah2021adversarial}. Neural representations
provide alternative parameterizations for adversarial objects. Adv3D
\cite{li2024adv3d} adopts neural radiance fields (NeRF)
\cite{mildenhall2020nerf} for attacks on camera-based 3D detection. Building
on 3DGS \cite{kerbl20233dgs}, PGA, 3DGAA, and BiTAA
study multi-view camouflage, appearance and geometry optimization, and
joint attacks on object detection and depth estimation, respectively
\cite{lou20253d,zhang2025dgaa,zhang2026bitaa}. These attacks primarily target
camera-based perception. Whereas 3DGS represents scenes with volumetric
Gaussian primitives, 2DGS uses oriented planar Gaussian disks that
explicitly encode local surface directions \cite{huang20242dgs}, providing
a surface-oriented basis for appearance rendering and geometry refinement.

Sensor simulation models how scene geometry and appearance produce
observations at different sensor poses.
LiDARsim combines ray casting with learned correction for LiDAR simulation
\cite{manivasagam2020lidarsim}, while SurfelGAN synthesizes camera views
from texture-mapped surfels \cite{yang2020surfelgan}. UniSim and NeuRAD
extend neural rendering to camera and LiDAR observations in dynamic driving
scenes \cite{yang2023unisim,tonderski2024neurad}. SplatAD further demonstrates
real-time rendering of both modalities using 3DGS, accounting for
sensor-specific effects such as rolling shutter and ray dropout
\cite{hess2025splatad}. These methods focus on scene reconstruction and
sensor simulation. AdvMTO uses a surface-oriented 2DGS representation for
bounded optimization of an adversarial object's appearance and geometry.

\section{Method}

\subsection{Problem Formulation}

We study white-box evasion of an inserted object against a frozen BEV
detector $F_d$. Each detector $d$ uses one adversarial object state across
observation records. Each record comprises a current sample, an object
placement, and the history consumed by the detector. The object's world pose
is fixed within that history and current sample but may vary between records.
For record $u\sim\mathcal U$, let $\mathbf x^{\mathrm{bg}}_{d,u}$ denote
recorded sensor inputs and $\zeta_u$ the target class and location.
A detector-specific constructor $\mathcal A_{d,u}$ forms observations from
a candidate object $\mathcal G$, yielding
\begin{equation}
\begin{aligned}
\mathbf x_{d,u}(\mathcal G)
&=\mathcal A_{d,u}\!\left(\mathbf x^{\mathrm{bg}}_{d,u},\mathcal G\right),\\
\mathcal G_d^\star
&\in\arg\min_{\mathcal G\in\Omega_d(\mathcal G_0)}
\mathbb E_{u\sim\mathcal U}
\Bigl[\ell_d\!\left(F_d(\mathbf x_{d,u}(\mathcal G)),\zeta_u\right)\Bigr].
\end{aligned}
\label{eq:attack-formulation}
\end{equation}
Here, $\mathcal G_0$ is the clean object, $\Omega_d$ bounds its admissible
changes, and $\ell_d$ is the detector-specific target-suppression loss.
Appearance is optimized
for all detectors, with geometry refinement for fusion detectors.
Detector parameters, calibration, ego poses, and timestamps remain fixed.
Figure~\ref{fig:framework} summarizes the framework.

\subsection{Surface-Oriented 2DGS Object Representation}
\label{sec:object-representation}

We reconstruct the clean object offline from posed multi-view RGB images
as $N$ oriented 2DGS surfels,
\begin{equation}
\mathcal G=\{g_i\}_{i=1}^{N},\qquad
g_i=(\boldsymbol\mu_i,\mathbf s_i,\mathbf q_i,\alpha_i,\mathbf c_i),
\label{eq:2dgs-state}
\end{equation}
with center $\boldsymbol\mu_i\in\mathbb R^3$, in-plane scale
$\mathbf s_i\in\mathbb R_+^2$, orientation $\mathbf q_i$, opacity
$\alpha_i\in[0,1]$, and view-independent RGB component
$\mathbf c_i\in[0,1]^3$. Orientation $\mathbf q_i$ defines a local plane
and its normal, which determines the displacement direction in geometry
refinement. The same surfels support camera rendering and LiDAR-view depth
and opacity queries.

Writing $\Theta=\{(\boldsymbol\mu_i,\mathbf s_i,\mathbf q_i)\}_{i=1}^N$,
$\boldsymbol\alpha=\{\alpha_i\}_{i=1}^N$, and
$\mathbf C=\{\mathbf c_i\}_{i=1}^N$, the clean state is
$\mathcal G_0=\mathcal G(\Theta_0,\boldsymbol\alpha_0,\mathbf C_0)$.
Attacks modify RGB components and, for geometry refinement, surfel centers.
Scales, orientations, opacities, and higher-order view-dependent appearance
coefficients remain fixed.

\subsection{Detector-Native Observation Construction}
\label{sec:observation-construction}

We generate current and historical observations of the candidate object
under recorded sensor poses. Native temporal processing then aligns LiDAR
sweeps for fusion detectors and updates BEV memory for temporal camera
detectors.

\paragraph{Object Placement and Sensor Poses.}
Let $t$ denote the current observation, $s$ any consumed observation, and
$T_{A\leftarrow B}$ the transform from frame $B$ to $A$. We use $O$, $W$,
and $E_s$ for object, world, and ego frames. After fixed scaling and ground
alignment, placement in reference camera $C_{p,t}$ determines the world
pose $T_{W\leftarrow O}^{(u)}=
T_{W\leftarrow E_t}T_{E_t\leftarrow C_{p,t}}T_{C_{p,t}\leftarrow O}^{(u)}$.
For camera $C_{v,s}$ or native LiDAR frame $L_s$, the object pose is
\begin{equation}
T_{S_s\leftarrow O}^{(u)}
=T_{S_s\leftarrow E_s}T_{E_s\leftarrow W}T_{W\leftarrow O}^{(u)},
\label{eq:sensor-object-pose}
\end{equation}
where $S_s\in\{C_{v,s},L_s\}$. Recorded calibration and ego poses determine
these transforms while the object's world pose remains fixed.

\paragraph{Camera Observation Rendering.}
For selected camera observation $(v,s)$, intrinsics $\mathbf K_{v,s}$ and
object pose determine the premultiplied RGB image $\mathbf B^{\mathcal G}_{v,s}$
and opacity $\mathbf A^{\mathcal G}_{v,s}$:
\begin{equation}
\begin{aligned}
(\mathbf B^{\mathcal G}_{v,s},\mathbf A^{\mathcal G}_{v,s})
&=\mathcal R_{\mathrm{cam}}\!\left(
\mathcal G;\mathbf K_{v,s},T_{C_{v,s}\leftarrow O}^{(u)}\right),\\
\widetilde{\mathbf I}_{v,s}(\mathcal G)
&=\bar{\mathbf A}^{\mathcal G}_{v,s}\odot\mathbf B^{\mathcal G}_{v,s}
+(1-\bar{\mathbf A}^{\mathcal G}_{v,s})\odot\mathbf I^{\mathrm{bg}}_{v,s}.
\end{aligned}
\label{eq:camera-render-composite}
\end{equation}
Here, $\bar{\mathbf A}^{\mathcal G}_{v,s}
=\operatorname{clip}(\mathbf A^{\mathcal G}_{v,s},0,1)$ and
$\mathbf I^{\mathrm{bg}}_{v,s}$ is the recorded image. Compositing precedes
detector preprocessing at the original resolution, retaining the
alpha-weighted renderer buffer used in the experiments. Unselected
observations remain unchanged.

\paragraph{Native-Sweep Scanline LiDAR Return Generation.}
Our generator queries 2DGS depth and opacity along prescribed LiDAR
scanlines independently for each current and historical source sweep.
As the ego vehicle moves, the scanlines intersect different surface regions
of the world-fixed object. Reusing current returns in historical sweeps
preserves the current sampling pattern even after coordinate alignment.
We therefore generate returns under each source pose before assembling the
temporal input.

An auxiliary raster frame $V_s$ shares the origin of $L_s$.
The fixed, zero-translation transform $T_{V\leftarrow L}$ rotates LiDAR
axes into the rasterizer's projection convention, giving
$T_{V_s\leftarrow O}^{(u)}=T_{V\leftarrow L}T_{L_s\leftarrow O}^{(u)}$.
With virtual intrinsics $\mathbf K_V$, we render axial depth $\mathbf D_s$
and accumulated opacity $\mathbf A_s$:
\begin{equation}
(\mathbf D_s,\mathbf A_s)=\mathcal R_{\mathrm{lidar}}\!\left(
\mathcal G;\mathbf K_V,T_{V_s\leftarrow O}^{(u)}\right).
\label{eq:lidar-buffers}
\end{equation}
We query these depth and opacity buffers along the prescribed LiDAR scanlines.
Each ring elevation and sampled azimuth defines a unit ray
$\mathbf r^{L_s}$, rotated as
$\mathbf r^{V_s}=\mathbf R_{V\leftarrow L}\mathbf r^{L_s}$.
We project it using $\pi_{\mathbf K_V}$ and bilinearly sample the depth
and opacity buffers:
\begin{equation}
\begin{aligned}
\mathbf q_s&=\pi_{\mathbf K_V}(\mathbf r^{V_s}),\\
(d_s,a_s)&=\operatorname{bilin}\bigl((\mathbf D_s,\mathbf A_s),\mathbf q_s\bigr).
\end{aligned}
\label{eq:scanline-query}
\end{equation}
We retain samples with valid projections and depths, sufficient opacity, and
$\mathbf e_z^\top\mathbf r^{V_s}>0$. Since $d_s$ is depth along the
raster's forward axis $\mathbf e_z=(0,0,1)^\top$, the return uses the
corresponding ray range:
\begin{equation}
\rho_s=\frac{d_s}{\mathbf e_z^\top\mathbf r^{V_s}}+\epsilon_s,
\qquad \mathbf p^{L_s}=\rho_s\mathbf r^{L_s}.
\label{eq:scanline-return}
\end{equation}
Dividing by the ray's forward component converts axial depth into distance
along the ray, placing the candidate in the original LiDAR frame.
Range perturbation $\epsilon_s$, dropout, point-count control, and intensity
assignment follow the fixed configuration in the supplement.

To account for occlusion between the object and the recorded scene, we
assign candidate returns and background points to common azimuth--elevation
cells of $L_s$. Within each cell, candidates lying behind
the nearest background return by more than a fixed range tolerance are
discarded. For each retained candidate, background points lying behind it
by more than the same tolerance are removed. The surviving object
and background returns are combined into one source cloud. This approximates
range-based visibility, yielding $\widehat{\mathcal P}_{u,s}(\mathcal G)$
in the source frame for subsequent temporal alignment.

\paragraph{Detector-Native Temporal Integration.}
For fusion detectors, the native input pipeline aligns each source cloud
$\widehat{\mathcal P}_{u,s}(\mathcal G)$ to the current LiDAR frame:
\begin{equation}
T_{L_t\leftarrow L_s}=T_{L_t\leftarrow W}T_{W\leftarrow L_s},
\qquad \Delta\tau_s=\tau_t-\tau_s,
\label{eq:temporal-lidar-alignment}
\end{equation}
where timestamps are expressed in seconds. Points are transformed once,
retain their intensities, and receive time lag $\Delta\tau_s$ before the
sweeps are concatenated. The aligned clouds retain source-dependent sampling
and visibility established during return generation.

For temporal camera detectors, we reconstruct each record's chronological
same-scene prefix $s_1,\ldots,s_m,t$ using one candidate object state and
fixed world pose. We reset the temporal state for each record, so different
placements reconstruct their own histories even when they share a current
frame. Let $\widetilde{\mathbf X}_{u,s}(\mathcal G)$ denote the camera inputs,
with the object composited into selected views and recorded images used
for the remaining views. Starting from
the detector-defined initial memory $\mathbf h_0$, native historical and
current computations give
\begin{equation}
\begin{aligned}
\mathbf h_j&=\Phi_{\mathrm{hist}}\!\left(
\mathbf h_{j-1},\widetilde{\mathbf X}_{u,s_j}(\mathcal G),
\boldsymbol\xi_{s_j}\right),\\
\mathbf y_t&=\Phi_{\mathrm{cur}}\!\left(
\mathbf h_m,\widetilde{\mathbf X}_{u,t}(\mathcal G),\boldsymbol\xi_t\right),
\end{aligned}
\label{eq:temporal-camera-memory}
\end{equation}
Here, $\Phi_{\mathrm{hist}}$ updates BEV memory for $j=1,\ldots,m$,
$\Phi_{\mathrm{cur}}$ computes the current prediction, and
$\boldsymbol\xi_s$ contains recorded pose and temporal metadata. Historical
frames without an object rendering still participate in the memory updates.

We evaluate the target-suppression loss at the current timestamp and
backpropagate through the current rendering and every historical BEV-memory
update. Gradients from current and historical renderings jointly optimize
the same object appearance, while detector weights remain frozen.

\subsection{Adversarial Optimization}
\label{sec:optimization}

\paragraph{Bounded appearance optimization.}
During appearance optimization, object geometry remains fixed, as do the
LiDAR inputs of fusion detectors. We optimize a detector-specific RGB residual
$\boldsymbol\delta_d$ on editable surfels:
\begin{equation}
\begin{aligned}
\mathbf C_d(\boldsymbol\delta_d)
&=\mathbf C_0+\mathbf M_d\odot\boldsymbol\delta_d,\\
\|\boldsymbol\delta_d\|_\infty&\leq\epsilon_d,
\qquad 0\leq\mathbf C_d(\boldsymbol\delta_d)\leq1.
\end{aligned}
\label{eq:bounded-appearance}
\end{equation}
The binary mask $\mathbf M_d$ identifies editable surfels. After each
update, we zero the residual outside this support and project each editable
entry onto $[-\epsilon_d,\epsilon_d]\cap[-\mathbf C_0,1-\mathbf C_0]$,
with bounds applied elementwise. This enforces both the perturbation budget
and valid RGB values, giving
$\mathcal G_d(\boldsymbol\delta_d)=
\mathcal G(\Theta_0,\boldsymbol\alpha_0,\mathbf C_d(\boldsymbol\delta_d))$.
We suppress local traffic-cone heatmap responses for BEVFusion and the
target-associated proposal score for DeepInteraction. For BEVFormer, we
minimize a spatially weighted surrogate over traffic-cone query logits,
with gradients propagated through historical BEV memory.
We aggregate record gradients,
add palette, white-band chroma, and local-smoothness regularization, then
update and project the residual.
Appearance checkpoints are selected by validation TS. Exact losses,
aggregation, and schedules are provided in the supplement.

\paragraph{Bounded geometry refinement.}
For fusion detectors, we displace editable 2DGS centers along outward
unit normals $\widehat{\mathbf n}_i$ derived from their fixed orientations:
\begin{equation}
\begin{aligned}
\boldsymbol\mu_i(\boldsymbol\eta)
&=\boldsymbol\mu_i^0+a\,\omega_i(\boldsymbol\eta)\widehat{\mathbf n}_i,\\
\max_i\|\boldsymbol\mu_i(\boldsymbol\eta)-\boldsymbol\mu_i^0\|_2
&\leq B.
\end{aligned}
\label{eq:bounded-geometry}
\end{equation}
Here, $\boldsymbol\eta$ specifies amplitude $a$ and spatial support.
Weights $\omega_i\in[0,1]$ localize the deformation and are zero on protected
centers. Each candidate regenerates camera images and native-sweep LiDAR
returns while all other object attributes remain fixed. Because observation
and detection include discrete operations, we select geometry from predefined
candidates on Validation. During geometry selection, we hold BEVFusion's
selected adversarial appearance fixed, while using the clean appearance
for DeepInteraction. For DeepInteraction, we then combine the selected
geometry with the selected adversarial appearance. Geometry is frozen
before Test for both detectors. Normal orientation, support weights, and detector-specific screening
and selection criteria are detailed in the supplement.

\section{Experiments}

We evaluate attack effectiveness across three BEV detectors and examine
the effects of LiDAR history construction and gradient propagation through
BEV memory. We also assess whether bounded geometry refinement improves
suppression over appearance optimization alone.\footnote{We used OpenAI ChatGPT and Codex for manuscript revision and review, code development and debugging, and analysis and verification of experimental results.}

\subsection{Experimental Setup}
\label{sec:experimental-setup}

\paragraph{Dataset.}
We conduct experiments on nuScenes~\cite{caesar2020nuscenes} using its
synchronized camera and LiDAR sequences. The main representation comparison
uses 919 object placements across 436 scenes, partitioned into disjoint
attack-training, validation, and test sets.
Detailed data splits and
clean-detectability screening criteria are provided in the supplementary
material.

\paragraph{Target models.}
We evaluate AdvMTO on three BEV-based 3D detectors.
BEVFusion~\cite{liu2023bevfusion} and
DeepInteraction~\cite{yang2022deepinteraction} consume six current camera
views and the current LiDAR sweep together with nine and ten historical
sweeps, respectively. BEVFormer~\cite{li2022bevformer} aggregates six-view
camera observations over a variable-length temporal prefix through BEV
memory. All detector parameters are frozen, and adversarial object states
are optimized separately for each detector. For camera inputs, object
rendering and appearance optimization use CAM\_FRONT, while the other
views remain unchanged.

\paragraph{Evaluation metrics.}
We measure attack effectiveness using Target Score (TS), relative score
reduction, and attack success rate (ASR). TS is the maximum post-processed
traffic-cone confidence among detections whose BEV centers lie within
1\,m of the inserted-object center, and is set to zero if no matching
detection is returned. Relative reduction is computed as
$(\mathrm{clean\ TS}-\mathrm{adversarial\ TS})/\mathrm{clean\ TS}$.
For the method, representation, and historical-gradient comparisons, ASR
measures the fraction of clean-detectable placements (TS $\geq 0.3$) whose adversarial TS
falls below 0.3.
We quantify appearance distortion using RGB MAE over object support and
LPIPS/SSIM over matched local crops.
Further details are provided in the supplement.

\subsection{Attack and Representation Comparisons}
\label{sec:main-appearance-results}

\input{tables/method_comparison}

\begin{figure*}[!t]
\centering
\includegraphics[width=\textwidth]{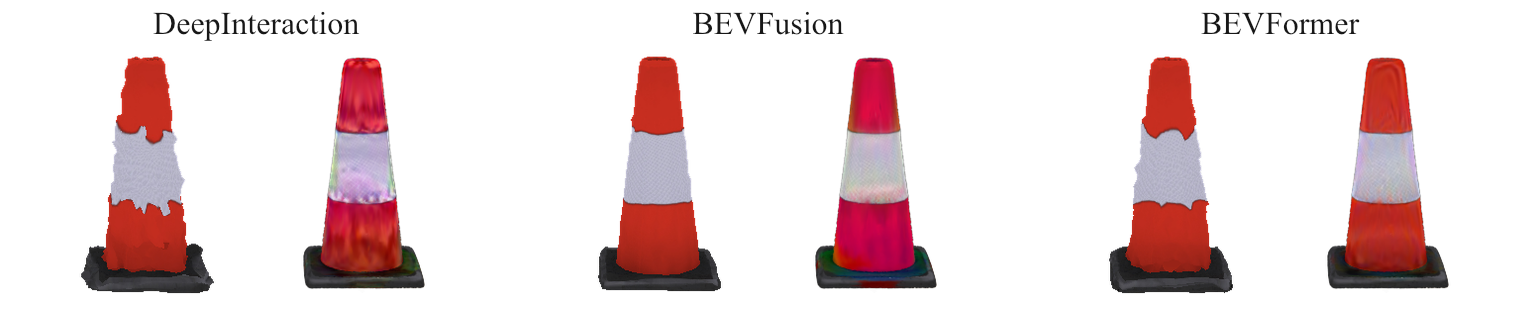}
\caption{Visual comparison of detector-specific adversarial objects.
Each pair shows MSF-ADV$^{\dagger}$ with geometry optimization and a fixed
texture (left) and AdvMTO with appearance-only optimization (right).
All objects are shown from the same frontal viewpoint. $^{\dagger}$ denotes
our implementation adapted to the evaluated detectors.}
\label{fig:detector-specific-states}
\end{figure*}

\input{tables/main_results}

\paragraph{Comparison with MSF-ADV.}
Table~\ref{tab:method-comparison} compares AdvMTO with our implementation of
MSF-ADV~\cite{cao2021lidaradvmulti}, adapted to optimize one universal mesh
with fixed texture per detector. The meshes are optimized on the
attack-training split and selected on validation, with implementation
details provided in the supplementary material.

The clean 2DGS object receives higher TS than the clean MSF-ADV mesh on
DeepInteraction and BEVFormer, while the reverse holds on BEVFusion.
After optimization, AdvMTO appearance attacks achieve lower adversarial TS
and higher ASR@0.3 than MSF-ADV on all three detectors while preserving
the original object geometry. Figure~\ref{fig:detector-specific-states}
shows visible surface distortions in the MSF-ADV meshes, particularly
around the white band of the DeepInteraction object.

For geometry-only attacks, MSF-ADV yields lower TS on DeepInteraction,
while AdvMTO yields lower TS on BEVFusion. Combining appearance and
geometry in AdvMTO achieves the lowest TS on both fusion detectors.
On BEVFusion, adding geometry refinement to the appearance attack
increases ASR@0.3 from 22.89\% to 80.72\%. These results show that AdvMTO
achieves stronger suppression through appearance optimization across the
evaluated detectors, with geometry refinement providing further gains
on the fusion models.

\begin{figure*}[!t]
\centering
\includegraphics[width=0.95\textwidth]{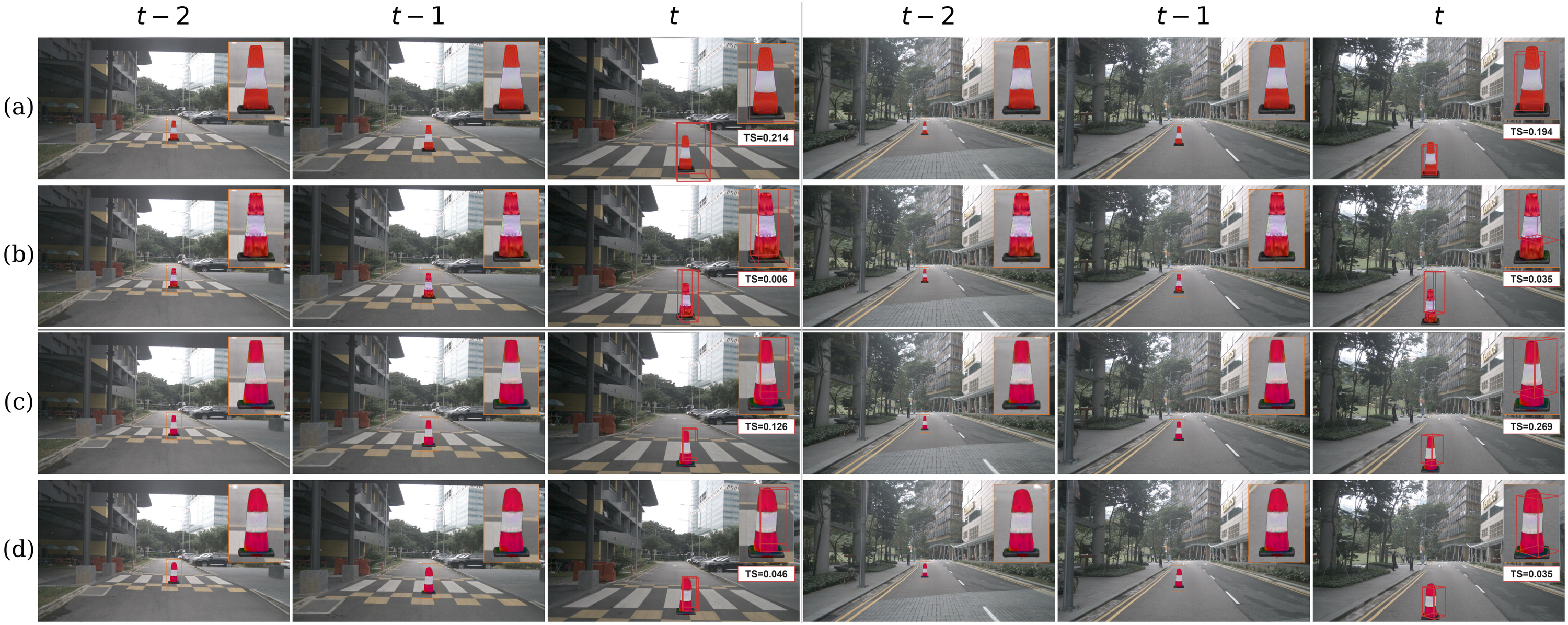}
\caption{Temporal views of adversarial objects in two test scenes.
Rows show appearance attacks on (a) BEVFormer, (b) DeepInteraction,
and (c) BEVFusion, followed by (d) BEVFusion with geometry refinement.
Current-frame panels report TS. Historical RGB views illustrate object
persistence for the fusion models, which consume historical LiDAR sweeps.}
\label{fig:qualitative-results}
\end{figure*}

\paragraph{Representation comparison.}
Table~\ref{tab:main-results} compares appearance attacks using 2DGS and
3DGS with fixed object geometry. Across all three detectors, the clean
2DGS object receives higher TS, and appearance optimization produces
larger relative TS reductions. The advantage is most pronounced on
BEVFormer, where the relative TS reduction increases from 59.5\% to 87.0\%
and ASR@0.3 increases by 32.6 percentage points. This improvement is
accompanied by lower LPIPS but higher MAE. On DeepInteraction, both
representations achieve 100\% ASR@0.3, while 2DGS further reduces
adversarial TS at higher MAE and LPIPS. On BEVFusion, 2DGS achieves a
larger relative TS reduction and higher ASR@0.3 at similar MAE and LPIPS,
while 3DGS retains a lower absolute adversarial TS.
The combination of higher clean detection confidence and larger relative
score reductions supports the use of 2DGS as the object representation in
AdvMTO. Figure~\ref{fig:qualitative-results} illustrates AdvMTO objects across
successive viewpoints in two test scenes, with all displayed current-frame
TS values below 0.3.

\subsection{Detector-Native Temporal Construction}
\label{sec:construction-diagnostics}

We examine how LiDAR history construction affects attack evaluation and
how gradient propagation through BEV memory affects attack optimization.

\paragraph{Temporal LiDAR construction.}
Table~\ref{tab:temporal-construction} compares LiDAR history constructions
on BEVFusion using fixed clean and appearance-plus-geometry (AG) objects.
Current-only inserts object returns into the current sweep.
Repeated-current copies them into historical sweeps, while native per-sweep
construction regenerates them at each source pose before temporal alignment.

\begin{table}[!htbp]
  \caption{Effect of temporal LiDAR construction on BEVFusion. $\Delta\mathrm{TS}$
  denotes the clean-to-adversarial score decrease.}
  \label{tab:temporal-construction}
  \centering
  \small
  \setlength{\tabcolsep}{1.9pt}
  \begin{tabular*}{\columnwidth}{@{\extracolsep{\fill}}lcccc@{}}
  \toprule
  Construction
  & Clean TS
  & Adv.\ TS $\downarrow$
  & $\Delta$TS $\uparrow$
  & ASR@0.3 $\uparrow$ \\
  \midrule
  Current-only     & 0.3267 & 0.1912 & 0.1355 & 77.90\% \\
  Repeated-current & 0.3949 & 0.2459 & 0.1490 & 58.56\% \\
  Native per-sweep & 0.4434 & \textbf{0.1802} & \textbf{0.2632}
  & \textbf{82.32\%} \\
  \bottomrule
  \end{tabular*}
\end{table}

Compared with repeated-current, native per-sweep construction reduces
adversarial TS by 26.7\% and increases ASR@0.3 by 23.8 percentage points.
Current-only produces a similar adversarial TS to native per-sweep
construction, while lowering clean TS from 0.4434 to 0.3267.
Its clean-to-adversarial score decrease is
consequently 0.1355, compared with 0.2632 for native per-sweep construction.
These results show that LiDAR history construction changes both the object's
clean detectability and the attack effect measured against that baseline.

\begin{table}[!htbp]
  \caption{Effect of historical gradient propagation on BEVFormer.
  The current frame is adversarial in both test conditions.}
\label{tab:bevformer-temporal-ablation}
  \centering
  \small
  \setlength{\tabcolsep}{2.5pt}
  \begin{tabular*}{\columnwidth}{@{\extracolsep{\fill}}lcccc@{}}
  \toprule
  & \multicolumn{2}{c}{Clean-object history}
  & \multicolumn{2}{c}{Adv.\ history} \\
  \cmidrule(lr){2-3}
  \cmidrule(lr){4-5}
  Optimization
  & TS $\downarrow$ & ASR@0.3 $\uparrow$
  & TS $\downarrow$ & ASR@0.3 $\uparrow$ \\
  \midrule
  Stop-gradient
  & \textbf{0.2291} & \textbf{72.93\%} & 0.1253 & 83.98\% \\
  Full-prefix
  & 0.2390 & 71.27\% & \textbf{0.1029} & \textbf{85.64\%} \\
  \bottomrule
  \end{tabular*}
\end{table}

\input{tables/pose_generalization}
\input{tables/geometry_four_states}

\paragraph{Historical gradient propagation.}
Table~\ref{tab:bevformer-temporal-ablation} evaluates historical gradient
propagation in BEVFormer. Both variants train with adversarial-object
histories under the same appearance budget and optimization schedule.
Full-prefix optimization backpropagates the current detection loss through
historical BEV memory and renderings, while stop-gradient detaches the
historical memory before current-frame prediction. At test time, the current
object is adversarial, while historical observations contain either its
clean or adversarial appearance.

Under adversarial histories, full-prefix optimization lowers TS from
0.1253 to 0.1029 relative to stop-gradient, a 17.9\% reduction. With
clean-object histories, the ranking reverses: stop-gradient yields a TS
of 0.2291, compared with 0.2390 for full-prefix optimization. For each
frozen attack state, replacing clean histories with adversarial ones
lowers TS without changing the current-frame image, indicating that
historical observations contribute to current-frame suppression. The
additional full-prefix gain under adversarial histories shows that
backpropagation through BEV memory strengthens suppression of a persistent
adversarial object.

\begin{figure}[!htbp]
\centering
\includegraphics[width=\columnwidth]{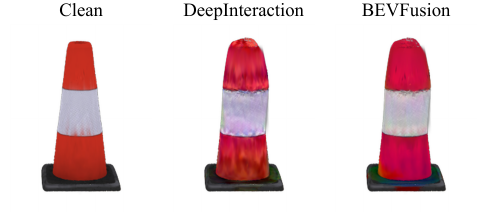}
\caption{Clean and AG 2DGS objects at matched view and scale.}
\label{fig:geometry-refinement}
\end{figure}

\subsection{Pose Generalization and Modality Ablations}
\label{sec:generalization-budget}
\label{sec:geometry-results}

\paragraph{Pose generalization.}
Multi-angle training improves suppression at untrained headings over
matched single-angle controls on all three detectors.
Table~\ref{tab:pose-generalization} shows low mean TS across tested poses
for DeepInteraction, limited TS variation across headings for BEVFusion,
and low mean TS under lateral shifts for BEVFormer.
BEVFormer shows weaker suppression at 10\,m and $\pm45^\circ$.
Its multi-angle fine-tuning uses a 6.5\,m current-frame anchor
and headings of $-30^\circ$, $0^\circ$, and $+30^\circ$.
Changes in projected scale and visible appearance may limit
the transfer of the optimized perturbations beyond these
training poses.

\paragraph{Geometry refinement.}
Table~\ref{tab:geometry} compares clean objects with appearance (A),
geometry (G), and combined perturbations (AG). G and AG use the same
selected deformation per detector and representation, with Gaussian-center
displacements bounded by 10\,cm and protected centers fixed.
Figure~\ref{fig:geometry-refinement} shows the clean and AG 2DGS objects.
Geometry perturbations alone reduce TS on both detectors for both
representations. They yield lower TS than appearance perturbations on
BEVFusion, while the reverse holds on DeepInteraction.
AG yields the lowest TS. Relative to A, the reductions on BEVFusion are
51.0\% for 2DGS and 49.1\% for 3DGS. The corresponding
reductions on DeepInteraction are 41.7\% and 45.9\%. A and AG reach
100\% ASR@0.3 on DeepInteraction, where AG further lowers TS.

\input{tables/crossmodal_contributions}

\paragraph{RGB--LiDAR contributions.}
We further examine modality contributions using the fixed clean and AG
2DGS objects from Table~\ref{tab:geometry}. The AG camera observations
include changes from both appearance and geometry, while the AG LiDAR
observations reflect the selected deformation. Starting from clean inputs at
each test placement, we replace RGB, LiDAR, or both with observations
from the AG object. When only one modality is replaced, the other remains
clean. RGB replacement changes only the front camera, while LiDAR
replacement includes the current and all historical sweeps. No object is
reoptimized for these comparisons. Table~\ref{tab:crossmodal} shows greater
TS reductions from LiDAR replacement on BEVFusion and from RGB replacement
on DeepInteraction. Replacing both modalities yields the lowest TS on
both detectors. Thus, both modalities contribute to suppression.

\section{Conclusion}

AdvMTO uses surface-oriented 2DGS to construct current and historical
camera and LiDAR observations of one adversarial object across sensor
poses and timestamps.
The 2DGS attacks achieve larger relative TS reductions than the 3DGS
attacks across the three evaluated detectors. AdvMTO appearance attacks
also achieve lower adversarial TS and higher ASR@0.3 than our MSF-ADV
implementation across all three detectors. Native LiDAR histories
change both clean detectability and measured attack effectiveness.
On BEVFormer, adversarial histories lower TS with the current image held
fixed, while backpropagation through BEV memory further strengthens
suppression. Combining appearance and geometry yields the lowest TS on
the two fusion detectors, with contributions from both RGB and LiDAR
observations. Future work will evaluate fabrication fidelity and attack
effectiveness under real camera--LiDAR sensing.

\label{lastcontentpage}

\FloatBarrier
{\small
\bibliographystyle{ieeenat_fullname}
\bibliography{references}
}

\clearpage
\appendix
\section*{Supplementary Material}
\setcounter{secnumdepth}{2}
\setcounter{table}{0}\renewcommand{\thetable}{S\arabic{table}}
\setcounter{figure}{0}\renewcommand{\thefigure}{S\arabic{figure}}
\setcounter{algorithm}{0}\renewcommand{\thealgorithm}{S\arabic{algorithm}}
\setcounter{equation}{0}\renewcommand{\theequation}{S\arabic{equation}}
\section{Test Population and Metrics}
\label{sec:supp-protocol}
\input{supplementary/evaluation_scope}
\input{supplementary/evaluation_protocol}
\input{supplementary/method_details}
\input{supplementary/comparison_evidence}
\input{supplementary/temporal_evidence}
\input{supplementary/pose_evidence}
\input{supplementary/modality_evidence}
\end{document}

%% file: tables/method_comparison.tex
\begin{table*}[!t]
\caption{Comparison with MSF-ADV. A and G denote appearance and geometry
perturbations, respectively. ASR@0.3 is pooled over placements with clean TS
$\geq0.3$ for both the mesh and 2DGS objects.
$^{\dagger}$ denotes our implementation adapted to the evaluated detectors.}
\label{tab:method-comparison}
\centering
\small
\setlength{\tabcolsep}{3.5pt}
\begin{tabular*}{\textwidth}{@{\extracolsep{\fill}}lllrrrr@{}}
\toprule
Detector & Method & Perturbation & Clean TS & Adv.\ TS $\downarrow$
& Rel.\ drop $\uparrow$ & ASR@0.3 $\uparrow$ \\
\midrule
DeepInteraction & MSF-ADV$^{\dagger}$ & G & 0.4616 & 0.1852 & 59.89\% & 72.54\% \\
                & AdvMTO & A  & 0.5429 & 0.0284 & 94.78\% & 100.00\% \\
                & AdvMTO & G  & 0.5429 & 0.3383 & 37.68\% & 33.80\% \\
                & AdvMTO & AG & 0.5429 & 0.0165 & 96.96\% & 100.00\% \\
\midrule
BEVFusion       & MSF-ADV$^{\dagger}$ & G & 0.4734 & 0.4720 & 0.29\% & 3.01\% \\
                & AdvMTO & A  & 0.4434 & 0.3676 & 17.11\% & 22.89\% \\
                & AdvMTO & G  & 0.4434 & 0.2434 & 45.11\% & 60.24\% \\
                & AdvMTO & AG & 0.4434 & 0.1801 & 59.39\% & 80.72\% \\
\midrule
BEVFormer       & MSF-ADV$^{\dagger}$ & G & 0.7769 & 0.4234 & 45.50\% & 41.90\% \\
                & AdvMTO & A  & 0.7933 & 0.1029 & 87.03\% & 85.47\% \\
\bottomrule
\end{tabular*}
\end{table*}

%% file: tables/main_results.tex
\begin{table*}[!t]
\caption{Appearance attacks with 2DGS and 3DGS.
We use $\epsilon=0.40$ for DeepInteraction and BEVFusion and
$\epsilon=0.08$ for BEVFormer. ASR@0.3 is pooled over placements with clean
TS $\geq0.3$ under both representations.}
\label{tab:main-results}
\centering
\small
\setlength{\tabcolsep}{3.5pt}
\begin{tabular*}{\textwidth}{@{\extracolsep{\fill}}llrrrrrr@{}}
\toprule
Detector & Representation & Clean TS & Adv.\ TS $\downarrow$
& Rel.\ drop $\uparrow$ & ASR@0.3 $\uparrow$
& MAE $\downarrow$ & LPIPS $\downarrow$ \\
\midrule
DeepInteraction & 2DGS & 0.5429 & 0.0284 & 94.78\% & 100.00\% & 0.0522 & 0.0670 \\
                & 3DGS & 0.5322 & 0.0402 & 92.46\% & 100.00\% & 0.0487 & 0.0542 \\
\midrule
BEVFusion       & 2DGS & 0.4434 & 0.3676 & 17.11\% & 21.13\% & 0.0603 & 0.0738 \\
                & 3DGS & 0.4085 & 0.3464 & 15.19\% & 13.38\% & 0.0606 & 0.0734 \\
\midrule
BEVFormer       & 2DGS & 0.7933 & 0.1029 & 87.03\% & 85.14\% & 0.02148 & 0.01770 \\
                & 3DGS & 0.7544 & 0.3053 & 59.53\% & 52.57\% & 0.01781 & 0.02513 \\
\bottomrule
\end{tabular*}
\end{table*}

%% file: tables/pose_generalization.tex
\begin{table*}[!t]
\caption{Pose and position generalization of frozen multi-angle-trained
2DGS objects.}
\label{tab:pose-generalization}
\centering
\small
\setlength{\tabcolsep}{2pt}
\begin{tabular*}{\textwidth}{@{\extracolsep{\fill}}l*{14}{r}@{}}
\toprule
TS $\downarrow$ & \multicolumn{4}{c}{Distance (m)}
& \multicolumn{7}{c}{Heading angle ($^\circ$)}
& \multicolumn{3}{c}{Lateral offset (m)} \\
\cmidrule(lr){2-5}\cmidrule(lr){6-12}\cmidrule(lr){13-15}
Detector & 5 & 6.5 & 8 & 10
& $-45$ & $-30$ & $-15$ & 0 & $+15$ & $+30$ & $+45$
& $-1$ & 0 & $+1$ \\
\midrule
DeepInteraction & 0.0505 & 0.0066 & 0.0068 & 0.0089 & 0.0415 & 0.0063 & 0.0066 & 0.0066 & 0.0065 & 0.0083 & 0.0273 & 0.0088 & 0.0066 & 0.0120 \\
BEVFusion & 0.3473 & 0.2920 & 0.2097 & 0.1674 & 0.3006 & 0.2976 & 0.3022 & 0.2920 & 0.2835 & 0.3028 & 0.3114 & 0.2579 & 0.2920 & 0.3132 \\
BEVFormer & 0.1465 & 0.0676 & 0.2786 & 0.5964 & 0.5234 & 0.1564 & 0.2382 & 0.0676 & 0.1767 & 0.2141 & 0.5397 & 0.1290 & 0.0676 & 0.1791 \\
\bottomrule
\end{tabular*}
\end{table*}

%% file: tables/geometry_four_states.tex
\begin{table}[!htbp]
\caption{Geometry refinement across representations.}
\label{tab:geometry}
\centering
\small
\setlength{\tabcolsep}{3.5pt}
\begin{tabular*}{\columnwidth}{@{\extracolsep{\fill}}lrrrr@{}}
\toprule
TS $\downarrow$ & \multicolumn{2}{c}{BEVFusion}
& \multicolumn{2}{c}{DeepInteraction} \\
\cmidrule(lr){2-3}\cmidrule(lr){4-5}
State & 2DGS & 3DGS & 2DGS & 3DGS \\
\midrule
Clean & 0.4434 & 0.4085 & 0.5429 & 0.5322 \\
A & 0.3676 & 0.3464 & 0.0284 & 0.0402 \\
G & 0.2434 & 0.2282 & 0.3383 & 0.3295 \\
AG & \textbf{0.1801} & \textbf{0.1762} & \textbf{0.0165} & \textbf{0.0217} \\
\bottomrule
\end{tabular*}
\end{table}

%% file: tables/crossmodal_contributions.tex
\begin{table}[!htbp]
\caption{RGB--LiDAR contributions using fixed 2DGS objects.
AG denotes appearance-and-geometry adversarial observations.}
\label{tab:crossmodal}
\centering
\small
\setlength{\tabcolsep}{3.5pt}
\begin{tabular*}{\columnwidth}{@{\extracolsep{\fill}}llrr@{}}
\toprule
RGB & LiDAR & BEVFusion TS $\downarrow$ & DI TS $\downarrow$ \\
\midrule
Clean & Clean & 0.4434 & 0.5429 \\
AG & Clean & 0.3717 & 0.0491 \\
Clean & AG & 0.2411 & 0.3658 \\
AG & AG & \textbf{0.1801} & \textbf{0.0165} \\
\bottomrule
\end{tabular*}
\end{table}

%% file: supplementary/evaluation_scope.tex
\label{sec:supp-reproducibility}
We evaluate one cone design in held-out nuScenes recordings, digitally
inserting a version fitted separately for each detector.

%% file: supplementary/evaluation_protocol.tex
\paragraph{Scene selection.}
We split the 700 official nuScenes training scenes by log, stratifying by
location and using a fixed hash order. The 150 official validation scenes
form our Test set. In each scene we select the lower-median timestamp
available to all three detectors. We place the clean 2DGS cone 6.5\,m
ahead at each of three lateral offsets ($-0.25$, 0, $+0.25$\,m).
Of 450 candidate Test placements, 181 pass the clean-detection threshold
on all three detectors (40.2\%, Table~\ref{tab:supp-population}).
The primary comparisons therefore describe placements where all three
detectors initially recognize the same cone.

\begin{table}[H]
\centering
\small
\setlength{\tabcolsep}{2.7pt}
\caption{Joint clean screening with the canonical 2DGS cone.
A placement qualifies when all three clean target scores reach 0.3.
A scene qualifies when at least one of its three placements does.}
\label{tab:supp-population}
\begin{tabular}{@{}lrrr@{}}
\toprule
Split & Source scenes & Eligible scenes & Eligible placements \\
\midrule
Train      & 539 & 265 & 564 \\
Validation & 161 &  85 & 174 \\
Test       & 150 &  86 & 181 \\
\midrule
Total      & 850 & 436 & 919 \\
\bottomrule
\end{tabular}
\end{table}

\paragraph{Units of analysis.}
Target Score (TS) is the highest post-processed cone confidence within
1\,m of the object's BEV center, or zero if there is no match.
We average TS first within scenes and then across scenes, giving equal
weight to a scene with one retained placement and a scene with three.
Relative reduction divides the clean-to-adversarial change in this mean
by the corresponding object's clean mean. ASR@0.3 is the fraction of
eligible placements with adversarial TS strictly below 0.3.
Tables~1--2 of the main paper require both clean
objects in each comparison to reach 0.3. Tables~3--4 use the original
181 placements, including all three LiDAR constructions in Table~3.

For paired 95\% percentile intervals we resample whole scenes 10,000
times, retaining their placements and conditions together. For ASR we
sum successes and eligible placements within each draw before dividing.
The intervals quantify variation across scenes for fixed object states.
The representation and geometry comparisons use one run per variant.

%% file: supplementary/method_details.tex
\section{Implementation Details}
\label{sec:supp-method}

\subsection{Sensor Observations}

\paragraph{Camera images.}
The compositor operates on the original image before resizing,
normalization, or padding. In main-text Eq.~4, the renderer's RGB buffer
already includes opacity. The implementation weights this buffer by the
clipped opacity when compositing it with the recorded background. The compositor uses projected
support without a scene-depth test.

\paragraph{LiDAR returns.}
LiDAR visibility uses recorded background ranges
(Table~\ref{tab:supp-lidar-model}).
Object intensity uses the same near-cone mean at all distances.
Background points retain their recorded intensities.

\begin{table}[!htbp]
\centering\small
\caption{LiDAR observation model. Range and intensity parameters apply
to the inserted object. The angular visibility test operates in each
source sweep before temporal alignment.}
\label{tab:supp-lidar-model}
\begin{tabularx}{\columnwidth}{@{}lX@{}}
\toprule
Quantity & Value in evaluation \\
\midrule
Virtual raster & $704\times256$ pixels, $70^\circ\times40^\circ$ field of view \\
Scanlines & 32 rings, elevations $-30.67^\circ$ to $10.67^\circ$,
azimuth spacing $0.333^\circ$ \\
Valid samples & Opacity $\geq0.95$, axial depth $(0.01,70)$\,m \\
Range noise & Standard deviation $0.02$\,m \\
Dropout & $0.06$ per scanline, $0.22$ within a scanline \\
Object returns & At most five per ring. Counts above 34 are
subsampled to 22--34. Counts below 22 are retained. \\
Intensity & Normal with mean 60 and standard deviation 30,
clipped to $[5,128]$ \\
Background visibility & $0.2^\circ$ angular cells, $0.1$\,m range tolerance \\
\bottomrule
\end{tabularx}
\end{table}

\paragraph{Camera history.}
BEVFormer prefixes span 17--22 frames from the same nuScenes scene. A historical image
contains the inserted cone when at least one projected support vertex
has positive depth and falls within the $1600\times900$ front view.
Other historical frames update memory with their recorded images.
We reset memory for each placement and score the current frame.

\subsection{Appearance and Validation}

\paragraph{Detector responses.}
Validation and Test use post-processed TS. The training scores differ by detector.
BEVFusion uses the mean of the five largest traffic-cone
heatmap probabilities in an $11\times11$ BEV window at the target.
DeepInteraction uses the score of the proposal associated with the local
post-processed cone response for that record. Its stored color offset uses the
zero-order spherical harmonic coefficient, so the reported albedo offset
is $0.28209$ times the change in that coefficient.

For BEVFormer the score averages a local and a distance-weighted response
over final-decoder cone logits. Let $z_j$ denote a logit, $d_j$ the BEV
distance between its predicted center and the target, and
$L_\tau(\{x_j\})=\tau\log\sum_j \exp(x_j/\tau)$ with $\tau=0.25$.
The distance-weighted response uses
$L_\tau(\{z_j+\log\max(w_j,10^{-12})\})$, where
$w_j=\exp[-d_j^2/(2\sigma^2)]$ and $\sigma=1$\,m.
The local response uses $L_\tau$ over logits within 2\,m of the target.
If that set is empty, both terms use the distance-weighted response.
Only the current frame contributes this score in either history condition.

\paragraph{Sampling and selection.}
Each primary appearance run comprises four passes over the 265 training
scenes. An update averages 53 distinct scenes with one placement per
scene visit. The three appearance penalties apply
once to this average. The palette penalty uses color clusters derived
from the clean cone. The chroma penalty applies to the reflective white
band when chroma exceeds 0.04. Local smoothness measures differences
in RGB residuals over a fixed eight-neighbor graph of the clean surface.
The cone base is protected. Table~\ref{tab:supp-color-terms} gives the
weights for these terms.

\begin{table}[!htbp]
\centering\small
\caption{Appearance regularization in the primary 2DGS comparison.
The last column lists the weight and type of color difference.}
\label{tab:supp-color-terms}
\setlength{\tabcolsep}{3.2pt}
\begin{tabular}{@{}lrrl@{}}
\toprule
Detector & Palette & White chroma & Smoothness \\
\midrule
DeepInteraction & 0.005 & 0.050 & 0.050, absolute \\
BEVFusion       & 0.020 & 0.050 & 0.050, squared \\
BEVFormer       & 0.020 & 0.050 & 0.050, squared \\
\bottomrule
\end{tabular}
\end{table}

The primary 2DGS appearance runs use Adam for 20 updates and 1,060
training visits. The learning rate is 0.010 for the two fusion
detectors and 0.005 for BEVFormer.
At updates 0, 5, 10, 15, and 20, each state is scored on all 174
Validation placements from 85 scenes. Minimum scene-averaged TS selects
the checkpoint, with ties resolved in favor of the earlier update.
All three runs selected update 20.

\subsection{Geometry Constraints}
\label{sec:supp-geometry}

The local surface direction in main-text Eq.~11 is oriented outward
using its projection onto the horizontal radial direction of each
surfel. The 3DGS comparison uses the shortest covariance axis as a
normal proxy with the same orientation rule. The spatial weight is the
product of two Gaussian weights centered on the selected azimuth and
normalized height. It vanishes outside the selected angular sector and
on the base. Candidate deformations change centers
within a 10\,cm bound and preserve scale, orientation, opacity,
and color.

Both representations use the same 12 candidates, crossing two displacement
amplitudes, three azimuth centers, and two angular widths at a fixed
height center. The coarse Validation set contains 24
placements from 12 scenes. The two candidates with the lowest TS at
each displacement amplitude advance to the full Validation set.
Full-set eligibility requires fixed base centers, a symmetric
center-set Chamfer distance of at most 0.10\,m, mean full-image SSIM of
at least 0.95, and mean full-image LPIPS of at most 0.10. The median
object-return count ratio relative to the clean reference must lie in
$[0.5,1.5]$ across placements, with a nonzero count at every placement.
The ratio counts object returns over the current and historical sweeps
together. The full-image checks compare AG with A for BEVFusion and
G with the clean object for DeepInteraction. A candidate must produce
valid observations at every placement to remain eligible.

Among eligible candidates, BEVFusion selects the lowest scene-averaged
AG score with adversarial appearance fixed. DeepInteraction selects the
lowest G score with clean appearance and combines the selected geometry
with adversarial appearance only after selection. In both detectors,
2DGS and 3DGS select the same nominal candidate, applied along each
representation's own normal directions.

%% file: supplementary/comparison_evidence.tex
\section{Baseline and Representation Comparisons}
\label{sec:supp-comparisons}

\begin{figure*}[!t]
\centering
\includegraphics[width=0.84\textwidth,trim=0 11 0 14,clip]{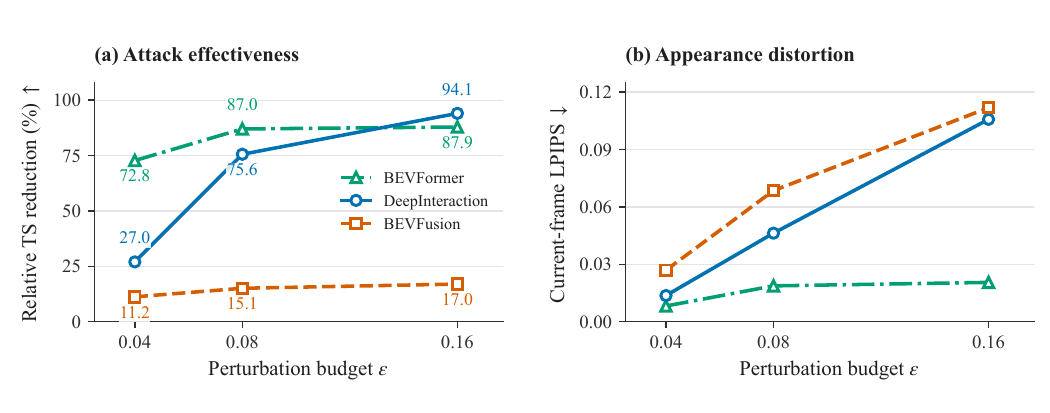}
\caption{2DGS budget comparison after 20 updates on 86 Test scenes.
Panel (a) reports TS reduction relative to each detector's clean mean.
Panel (b) measures LPIPS on detector-specific current-frame crops.
BEVFormer reuses the Table~2 checkpoint at $\epsilon=0.08$.}
\label{fig:supp-budget-cost}
\end{figure*}

\paragraph{MSF-ADV adaptation (main-text Table~1).}
\label{sec:supp-msf-comparison}
Table~1 evaluates our adaptation of scene-specific MSF-ADV to a single
fixed-texture mesh per detector across scenes. The mesh and AdvMTO
appearance runs each use 1,060 training-record visits. Different batch
sizes yield 133 mesh updates and 20 appearance updates.
MSF-ADV bounds each vertex
coordinate change at 0.20\,m. AdvMTO bounds the total displacement
of each Gaussian center at 0.10\,m.
Validation TS selects the mesh checkpoint separately for each detector.

\paragraph{Clean references (main-text Tables~1 and~2).}
For ASR, intersecting clean
detections with the mesh retains 142, 166, and 179 placements for
DeepInteraction, BEVFusion, and BEVFormer. Intersecting with 3DGS retains
171, 142, and 175. Thus the same BEVFormer appearance object has TS
0.1029 in both tables and ASR counts of 153/179 and 149/175.

\paragraph{Direct representation comparison.}
\label{sec:supp-representation}
The two clean reconstructions use the same 100 posed synthetic RGBA
views of one cone, yielding 15,819 surfels for 2DGS and 33,618 Gaussians
for 3DGS. Each representation uses its own color parameterization and,
on the fusion detectors, its own depth-derived LiDAR returns.
On BEVFusion, 2DGS starts 0.03494 higher in clean TS. Its relative
reduction is larger, yet its final TS is higher. The paired interval
for the latter difference spans zero (Table~\ref{tab:supp-representation-ci}).

\paragraph{Common appearance budget (main-text Table~2).}
Table~2 uses $\epsilon=0.40$ for both fusion detectors and $0.08$
for BEVFormer, with the same bound for both representations.
At a common 2DGS bound of $0.08$, relative TS reductions are 75.6\%
for DeepInteraction, 15.1\% for BEVFusion, and 87.0\% for BEVFormer
(Fig.~\ref{fig:supp-budget-cost}).

\begin{table}[!htbp]
\centering\small
\setlength{\tabcolsep}{3pt}
\caption{Paired differences (2DGS minus 3DGS) for main-text Table~2.
Each bootstrap draw uses the representation's own clean mean for
relative reduction. Differences in reduction are in percentage points.}
\label{tab:supp-representation-ci}
\begin{tabular}{@{}lrc@{}}
\toprule
Measure & Difference & 95\% CI \\
\midrule
\multicolumn{3}{@{}l}{\textit{DeepInteraction}} \\
Clean TS & $+0.01065$ & $[-0.00680, 0.02866]$ \\
Adversarial TS & $-0.01180$ & $[-0.01608,-0.00774]$ \\
Relative reduction (pp) & $+2.32$ & $[1.69,2.96]$ \\
\midrule
\multicolumn{3}{@{}l}{\textit{BEVFusion}} \\
Clean TS & $+0.03494$ & $[0.01335,0.05750]$ \\
Adversarial TS & $+0.02113$ & $[-0.00008,0.04288]$ \\
Relative reduction (pp) & $+1.92$ & $[0.70,3.09]$ \\
\midrule
\multicolumn{3}{@{}l}{\textit{BEVFormer}} \\
Clean TS & $+0.03889$ & $[0.01604,0.06830]$ \\
Adversarial TS & $-0.20249$ & $[-0.25348,-0.15167]$ \\
Relative reduction (pp) & $+27.51$ & $[21.08,33.93]$ \\
\bottomrule
\end{tabular}
\end{table}

\paragraph{Image costs and appearance controls.}
\label{sec:supp-random-controls}
Within each detector, we compare representations on the same image
regions. MAE averages over the union of the two clean object masks for
DeepInteraction and BEVFormer. For BEVFusion the union also includes
deformed objects. LPIPS and SSIM include background within each crop.

\begin{table}[!htbp]
\centering\small
\setlength{\tabcolsep}{3pt}
\caption{Random appearances at matched current-frame image cost.
For each detector we calibrated five random states to mean Test-frame
LPIPS using image distances alone, then froze them before scoring.
Random columns give the five-state means.}
\label{tab:random-appearance-controls}
\begin{tabular}{@{}lrrrr@{}}
\toprule
 & \multicolumn{2}{c}{TS} & \multicolumn{2}{c}{LPIPS} \\
\cmidrule(lr){2-3}\cmidrule(l){4-5}
Detector & Random & Optimized & Random & Optimized \\
\midrule
DeepInteraction & 0.5633 & 0.0284 & 0.06701 & 0.06704 \\
BEVFusion & 0.4498 & 0.3676 & 0.07374 & 0.07381 \\
BEVFormer & 0.8232 & 0.1029 & 0.01766 & 0.01770 \\
\bottomrule
\end{tabular}
\end{table}

The random states required larger parameter changes to match LPIPS.
Their mean TS exceeds that of the optimized object for each detector
(Table~\ref{tab:random-appearance-controls}).

%% file: supplementary/temporal_evidence.tex
\section{Temporal Ablations}
\label{sec:supp-temporal}

\paragraph{LiDAR history (main-text Table~3).}
Changing the history changes the clean score as well as the final
score. The native and current-only final
scores differ by $-0.0108$ with an interval spanning zero, despite the
larger clean-to-adversarial drop for native history. Against repeated-current,
both the final-score and ASR intervals exclude zero
(Table~\ref{tab:supp-temporal-pairs}).

\begin{table}[!htbp]
\centering
\caption{Paired differences for main-text Tables~3--4. TS gives equal
weight to each of 86 scenes. ASR pools all 181 placements and differences
are in percentage points. Intervals resample whole scenes.}
\label{tab:supp-temporal-pairs}
\small
\setlength{\tabcolsep}{3pt}
\begin{tabular*}{\columnwidth}{@{\extracolsep{\fill}}lrr@{}}
\toprule
Comparison & Final TS diff. & ASR diff. \\
\midrule
\multicolumn{3}{@{}l}{\textit{BEVFusion, native minus alternative}}\\
Repeated-current & $-0.0656$ & $+23.76$ \\
 & $(-0.0853,-0.0458)$ & $(+14.61,+32.61)$ \\
Current-only & $-0.0108$ & $+4.42$ \\
 & $(-0.0320,+0.0112)$ & $(-4.55,+13.26)$ \\
\midrule
\multicolumn{3}{@{}l}{\textit{BEVFormer, Full minus Stop}}\\
Adversarial history & $-0.0224$ & $+1.66$ \\
 & $(-0.0402,-0.0056)$ & $(-1.68,+5.46)$ \\
Clean-object history & $+0.0099$ & $-1.66$ \\
 & $(+0.0028,+0.0195)$ & $(-3.72,0.00)$ \\
\bottomrule
\end{tabular*}
\end{table}

\paragraph{Camera history (main-text Table~4).}
For each frozen BEVFormer object we keep the current image fixed when
switching Test histories. Full-prefix training gives lower TS than
stop-gradient under adversarial history and higher TS under clean-object
history. At the 0.3 threshold the corresponding success counts are
155 versus 152 and 129 versus 132. Both paired ASR intervals include
zero.

%% file: supplementary/pose_evidence.tex
\section{Pose Generalization}
\label{sec:supp-pose}
\label{sec:supp-generalization-budget}
\label{sec:supp-fusion-generalization}

\paragraph{Training and test poses (main-text Table~5).}
For all three detectors, angle fine-tuning uses a current-frame anchor
6.5\,m ahead with zero lateral offset. The single-angle condition uses
a heading of $0^\circ$. The multi-angle condition uses $-30^\circ$,
$0^\circ$, and $+30^\circ$. At Test we freeze each object and vary one
coordinate at a time. The $\pm15^\circ$ and $\pm45^\circ$ headings, distances of 5, 8,
and 10\,m, and lateral offsets of $\pm1$\,m were held out of fine-tuning.

\paragraph{Matched scenes.}
At each pose, both objects are scored on the same scenes that pass
visibility and occlusion checks, irrespective of clean TS. Eligible
scenes can differ across poses. The multi-angle object had lower TS
at the untrained $\pm45^\circ$ headings on all three detectors
(Table~\ref{tab:supp-fusion-angle-control}) and higher frontal TS.
BEVFormer also had higher TS at 10\,m (difference $+0.0925$).

\begin{table}[!htbp]
\centering\small
\caption{Multi-angle minus single-angle TS at held-out headings
(95\% scene-bootstrap intervals). The $\pm$ rows average both headings
per scene. DeepInteraction averages three paired seeds with
validation-selected epochs. Each other detector uses one seed and
equal training duration.}
\label{tab:supp-fusion-angle-control}
\label{tab:supp-former-generalization}
\setlength{\tabcolsep}{2pt}
\begin{tabular*}{\columnwidth}{@{\extracolsep{\fill}}lcrc@{}}
\toprule
Detector & Heading & $\Delta$TS & 95\% CI \\
\midrule
DeepInteraction & $\pm45^\circ$ & $-0.3699$ & $[-0.3892,-0.3509]$ \\
BEVFusion & $-45^\circ$ & $-0.0336$ & $[-0.0387,-0.0289]$ \\
           & $+45^\circ$ & $-0.0329$ & $[-0.0376,-0.0286]$ \\
BEVFormer & $\pm45^\circ$ & $-0.2758$ & $[-0.3929,-0.1708]$ \\
\bottomrule
\end{tabular*}
\end{table}

%% file: supplementary/modality_evidence.tex
\section{Geometry and Sensor Inputs}
\label{sec:supp-modalities}

\paragraph{Geometry at fixed appearance (main-text Table~6).}
Adding geometry lowers TS and increases crop LPIPS in all four
detector--representation pairs (Table~\ref{tab:supp-geometry-pairs}).
On BEVFusion, successes rise from 30 to 110 of 142 for 2DGS and from
19 to 112 of 142 for 3DGS. The symmetric Chamfer distances between
clean and deformed centers are 0.00757\,m and 0.00748\,m, respectively.
Main-text Table~3 uses an earlier frozen BEVFusion AG object.

\begin{table}[!htbp]
\centering
\caption{Geometry at fixed appearance (main-text Table~6).
$\Delta$TS is AG minus A on 181 placements. LPIPS uses each object's
clean image and a shared crop covering all clean and deformed supports.}
\label{tab:supp-geometry-pairs}
\small
\setlength{\tabcolsep}{2pt}
\begin{tabular*}{\columnwidth}{@{\extracolsep{\fill}}lrr@{}}
\toprule
Representation & $\Delta$TS & LPIPS (A / AG) \\
\midrule
\multicolumn{3}{@{}l}{\textit{DeepInteraction}}\\
2DGS & $-0.0118$ & 0.0657 / 0.0866 \\
3DGS & $-0.0184$ & 0.0533 / 0.0828 \\
\midrule
\multicolumn{3}{@{}l}{\textit{BEVFusion}}\\
2DGS & $-0.1875$ & 0.0738 / 0.0994 \\
3DGS & $-0.1702$ & 0.0734 / 0.1048 \\
\bottomrule
\end{tabular*}
\end{table}

\paragraph{Sensor substitution (main-text Table~7).}
AG images include the geometry change. On DeepInteraction, adding
LiDAR lowers TS even after RGB alone reaches 181/181 successes
(Table~\ref{tab:supp-modality-pairs}). On re-evaluation, one BEVFusion
clean score fell below 0.3, leaving 180 placements eligible for ASR.

\begin{table}[H]
\centering
\caption{Main-text Table~7. Paired TS drops on 181 placements and
success counts among clean detections.}
\label{tab:supp-modality-pairs}
\small
\setlength{\tabcolsep}{3pt}
\begin{tabular*}{\columnwidth}{@{\extracolsep{\fill}}lrr@{}}
\toprule
 & DeepInteraction & BEVFusion \\
\midrule
RGB given AG LiDAR & 0.34931 & 0.06108 \\
LiDAR given AG RGB & 0.03258 & 0.19159 \\
\midrule
AG RGB only & 181/181 & 44/180 \\
AG LiDAR only & 51/181 & 116/180 \\
Both AG & 181/181 & 148/180 \\
\bottomrule
\end{tabular*}
\end{table}

%% file: main.bbl
\begin{thebibliography}{39}
\providecommand{\natexlab}[1]{#1}
\providecommand{\url}[1]{\texttt{#1}}
\expandafter\ifx\csname urlstyle\endcsname\relax
  \providecommand{\doi}[1]{doi: #1}\else
  \providecommand{\doi}{doi: \begingroup \urlstyle{rm}\Url}\fi

\bibitem[Abdelfattah et~al.(2021)Abdelfattah, Yuan, Wang, and
  Ward]{abdelfattah2021adversarial}
Mazen Abdelfattah, Kaiwen Yuan, Z.~Jane Wang, and Rabab Ward.
\newblock Adversarial attacks on camera-{LiDAR} models for 3d car detection.
\newblock In \emph{2021 IEEE/RSJ International Conference on Intelligent Robots
  and Systems (IROS)}, pages 2189--2194. IEEE, 2021.

\bibitem[Caesar et~al.(2020)Caesar, Bankiti, Lang, Vora, Liong, Xu, Krishnan,
  Pan, Baldan, and Beijbom]{caesar2020nuscenes}
Holger Caesar, Varun Bankiti, Alex~H. Lang, Sourabh Vora, Venice~Erin Liong,
  Qiang Xu, Anush Krishnan, Yu Pan, Giancarlo Baldan, and Oscar Beijbom.
\newblock nuscenes: A multimodal dataset for autonomous driving.
\newblock In \emph{Proceedings of the IEEE/CVF Conference on Computer Vision
  and Pattern Recognition}, pages 11621--11631, 2020.

\bibitem[Cao et~al.(2019)Cao, Xiao, Cyr, Zhou, Park, Rampazzi, Chen, Fu, and
  Mao]{cao2019lidarattack}
Yulong Cao, Chaowei Xiao, Benjamin Cyr, Yimeng Zhou, Won Park, Sara Rampazzi,
  Qi~Alfred Chen, Kevin Fu, and Z.~Morley Mao.
\newblock Adversarial sensor attack on {LiDAR}-based perception in autonomous
  driving.
\newblock In \emph{Proceedings of the 2019 ACM SIGSAC Conference on Computer
  and Communications Security}, 2019.

\bibitem[Cao et~al.(2021)Cao, Wang, Xiao, Yang, Fang, Yang, Chen, Liu, and
  Li]{cao2021lidaradvmulti}
Yulong Cao, Ningfei Wang, Chaowei Xiao, Dawei Yang, Jin Fang, Ruigang Yang,
  Qi~Alfred Chen, Mingyan Liu, and Bo Li.
\newblock Invisible for both camera and {LiDAR}: Security of multi-sensor
  fusion based perception in autonomous driving under physical-world attacks.
\newblock In \emph{2021 IEEE Symposium on Security and Privacy (SP)}, pages
  176--194. IEEE, 2021.

\bibitem[Cao et~al.(2023)Cao, Bhupathiraju, Naghavi, Sugawara, Mao, and
  Rampazzi]{cao2023physicalremoval}
Yulong Cao, S.~Hrushikesh Bhupathiraju, Pirouz Naghavi, Takeshi Sugawara,
  Z.~Morley Mao, and Sara Rampazzi.
\newblock You can't see me: Physical removal attacks on {LiDAR}-based
  autonomous vehicles driving frameworks.
\newblock In \emph{32nd USENIX Security Symposium}, pages 2993--3010. USENIX
  Association, 2023.

\bibitem[Cheng et~al.(2022)Cheng, Liang, Choi, Tao, Cao, Liu, and
  Zhang]{cheng2022physicaldepth}
Zhiyuan Cheng, James Liang, Hongjun Choi, Guanhong Tao, Zhiwen Cao, Dongfang
  Liu, and Xiangyu Zhang.
\newblock Physical attack on monocular depth estimation with optimal
  adversarial patches.
\newblock In \emph{Computer Vision--ECCV 2022}, pages 514--532. Springer Nature
  Switzerland, 2022.

\bibitem[Cheng et~al.(2024)Cheng, Choi, Feng, Liang, Tao, Liu, Zuzak, and
  Zhang]{cheng2024fusion}
Zhiyuan Cheng, Hongjun Choi, Shiwei Feng, James~Chenhao Liang, Guanhong Tao,
  Dongfang Liu, Michael Zuzak, and Xiangyu Zhang.
\newblock Fusion is not enough: Single modal attacks on fusion models for 3d
  object detection.
\newblock In \emph{The Twelfth International Conference on Learning
  Representations}, 2024.

\bibitem[Hess et~al.(2025)Hess, Lindstr{\"o}m, Fatemi, Petersson, and
  Svensson]{hess2025splatad}
Georg Hess, Carl Lindstr{\"o}m, Maryam Fatemi, Christoffer Petersson, and
  Lennart Svensson.
\newblock {SplatAD}: Real-time {LiDAR} and camera rendering with 3d gaussian
  splatting for autonomous driving.
\newblock In \emph{Proceedings of the IEEE/CVF Conference on Computer Vision
  and Pattern Recognition}, pages 11982--11992, 2025.

\bibitem[Huang et~al.(2024)Huang, Yu, Chen, Geiger, and Gao]{huang20242dgs}
Binbin Huang, Zehao Yu, Anpei Chen, Andreas Geiger, and Shenghua Gao.
\newblock 2d gaussian splatting for geometrically accurate radiance fields.
\newblock In \emph{ACM SIGGRAPH 2024 Conference Papers}. Association for
  Computing Machinery, 2024.

\bibitem[Kerbl et~al.(2023)Kerbl, Kopanas, Leimkuehler, and
  Drettakis]{kerbl20233dgs}
Bernhard Kerbl, Georgios Kopanas, Thomas Leimkuehler, and George Drettakis.
\newblock 3d gaussian splatting for real-time radiance field rendering.
\newblock \emph{ACM Transactions on Graphics}, 42\penalty0 (4):\penalty0 1--14,
  2023.

\bibitem[Li et~al.(2026)Li, Xiang, Hou, Wan, Zhang, and Dai]{li2026saber}
Aixuan Li, Mochu Xiang, Bosen Hou, Zhexiong Wan, Jing Zhang, and Yuchao Dai.
\newblock Saber: Spatially consistent 3d universal adversarial objects for bev
  detectors.
\newblock In \emph{Proceedings of the IEEE/CVF Conference on Computer Vision
  and Pattern Recognition (CVPR)}, pages 25841--25850, 2026.

\bibitem[Li et~al.(2024)Li, Lian, and Chen]{li2024adv3d}
Leheng Li, Qing Lian, and Ying-Cong Chen.
\newblock Adv3d: Generating 3d adversarial examples for 3d object detection in
  driving scenarios with {NeRF}.
\newblock In \emph{2024 IEEE/RSJ International Conference on Intelligent Robots
  and Systems (IROS)}, pages 10813--10820. IEEE, 2024.

\bibitem[Li et~al.(2022)Li, Wang, Li, Xie, Sima, Lu, Qiao, and
  Dai]{li2022bevformer}
Zhiqi Li, Wenhai Wang, Hongyang Li, Enze Xie, Chonghao Sima, Tong Lu, Yu Qiao,
  and Jifeng Dai.
\newblock Bevformer: Learning bird's-eye-view representation from multi-camera
  images via spatiotemporal transformers.
\newblock In \emph{Computer Vision -- ECCV 2022}, pages 1--18. Springer, 2022.

\bibitem[Liang et~al.(2025)Liang, Liang, Lou, Zhang, Li, Fan, and
  Cao]{liang2025grac}
Jiawei Liang, Siyuan Liang, Tianrui Lou, Ming Zhang, Wenjin Li, Dunqiu Fan, and
  Xiaochun Cao.
\newblock Gradient-reweighted adversarial camouflage for physical object
  detection evasion.
\newblock In \emph{Proceedings of the IEEE/CVF International Conference on
  Computer Vision}, pages 13880--13889, 2025.

\bibitem[Liu et~al.(2023)Liu, Tang, Amini, Yang, Mao, Rus, and
  Han]{liu2023bevfusion}
Zhijian Liu, Haotian Tang, Alexander Amini, Xinyu Yang, Huizi Mao, Daniela Rus,
  and Song Han.
\newblock Bevfusion: Multi-task multi-sensor fusion with unified bird's-eye
  view representation.
\newblock In \emph{2023 IEEE International Conference on Robotics and
  Automation (ICRA)}, pages 2774--2781. IEEE, 2023.

\bibitem[Lou et~al.(2025)Lou, Jia, Liang, Liang, Zhang, Xiao, and
  Cao]{lou20253d}
Tianrui Lou, Xiaojun Jia, Siyuan Liang, Jiawei Liang, Ming Zhang, Yanjun Xiao,
  and Xiaochun Cao.
\newblock 3d gaussian splatting driven multi-view robust physical adversarial
  camouflage generation.
\newblock In \emph{Proceedings of the IEEE/CVF International Conference on
  Computer Vision (ICCV)}, pages 28752--28762, 2025.

\bibitem[Manivasagam et~al.(2020)Manivasagam, Wang, Wong, Zeng, Sazanovich,
  Tan, Yang, Ma, and Urtasun]{manivasagam2020lidarsim}
Sivabalan Manivasagam, Shenlong Wang, Kelvin Wong, Wenyuan Zeng, Mikita
  Sazanovich, Shuhan Tan, Bin Yang, Wei-Chiu Ma, and Raquel Urtasun.
\newblock Lidarsim: Realistic lidar simulation by leveraging the real world.
\newblock In \emph{Proceedings of the IEEE/CVF Conference on Computer Vision
  and Pattern Recognition}, pages 11167--11176, 2020.

\bibitem[Mildenhall et~al.(2020)Mildenhall, Srinivasan, Tancik, Barron,
  Ramamoorthi, and Ng]{mildenhall2020nerf}
Ben Mildenhall, Pratul~P. Srinivasan, Matthew Tancik, Jonathan~T. Barron, Ravi
  Ramamoorthi, and Ren Ng.
\newblock {NeRF}: Representing scenes as neural radiance fields for view
  synthesis.
\newblock In \emph{Computer Vision -- ECCV 2020}, pages 405--421. Springer,
  2020.

\bibitem[Philion and Fidler(2020)]{philion2020lift}
Jonah Philion and Sanja Fidler.
\newblock Lift, splat, shoot: Encoding images from arbitrary camera rigs by
  implicitly unprojecting to 3d.
\newblock In \emph{Computer Vision -- ECCV 2020}, pages 194--210. Springer,
  2020.

\bibitem[Shahriar et~al.(2026)Shahriar, Al~Barat, Sundar, Zhang, Ramakrishnan,
  Hou, and Lou]{shahriar2026temporal}
Md~Hasan Shahriar, Md~Mohaimin Al~Barat, Harshavardhan Sundar, Ning Zhang,
  Naren Ramakrishnan, Y.~Thomas Hou, and Wenjing Lou.
\newblock Temporal misalignment attacks against multimodal perception in
  autonomous driving.
\newblock In \emph{2026 IEEE Conference on Secure and Trustworthy Machine
  Learning (SaTML)}, 2026.

\bibitem[Suryanto et~al.(2022)Suryanto, Kim, Kang, Larasati, Yun, Le, Yang, Oh,
  and Kim]{suryanto2022dta}
Naufal Suryanto, Yongsu Kim, Hyoeun Kang, Harashta~Tatimma Larasati, Youngyeo
  Yun, Thi-Thu-Huong Le, Hunmin Yang, Se-Yoon Oh, and Howon Kim.
\newblock Dta: Physical camouflage attacks using differentiable transformation
  network.
\newblock In \emph{Proceedings of the IEEE/CVF Conference on Computer Vision
  and Pattern Recognition}, pages 15305--15314, 2022.

\bibitem[Suryanto et~al.(2023)Suryanto, Kim, Larasati, Kang, Le, Hong, Yang,
  Oh, and Kim]{suryanto2023active}
Naufal Suryanto, Yongsu Kim, Harashta~Tatimma Larasati, Hyoeun Kang,
  Thi-Thu-Huong Le, Yoonyoung Hong, Hunmin Yang, Se-Yoon Oh, and Howon Kim.
\newblock {ACTIVE}: Towards highly transferable 3d physical camouflage for
  universal and robust vehicle evasion.
\newblock In \emph{Proceedings of the IEEE/CVF International Conference on
  Computer Vision}, pages 4305--4314, 2023.

\bibitem[Tonderski et~al.(2024)Tonderski, Lindstr{\"o}m, Hess, Ljungbergh,
  Svensson, and Petersson]{tonderski2024neurad}
Adam Tonderski, Carl Lindstr{\"o}m, Georg Hess, William Ljungbergh, Lennart
  Svensson, and Christoffer Petersson.
\newblock {NeuRAD}: Neural rendering for autonomous driving.
\newblock In \emph{Proceedings of the IEEE/CVF Conference on Computer Vision
  and Pattern Recognition}, pages 14895--14904, 2024.

\bibitem[Tu et~al.(2020)Tu, Ren, Manivasagam, Liang, Yang, Du, Cheng, and
  Urtasun]{tu2020physically}
James Tu, Mengye Ren, Sivabalan Manivasagam, Ming Liang, Bin Yang, Richard Du,
  Frank Cheng, and Raquel Urtasun.
\newblock Physically realizable adversarial examples for lidar object
  detection.
\newblock In \emph{Proceedings of the IEEE/CVF Conference on Computer Vision
  and Pattern Recognition}, pages 13716--13725, 2020.

\bibitem[Tu et~al.(2022)Tu, Li, Yan, Ren, Chen, Liang, Bitar, Yumer, and
  Urtasun]{tu2022multisensor}
James Tu, Huichen Li, Xinchen Yan, Mengye Ren, Yun Chen, Ming Liang, Eilyan
  Bitar, Ersin Yumer, and Raquel Urtasun.
\newblock Exploring adversarial robustness of multi-sensor perception systems
  in self driving.
\newblock In \emph{Proceedings of the 5th Conference on Robot Learning}, pages
  1013--1024. PMLR, 2022.

\bibitem[Wang et~al.(2022)Wang, Jiang, Sun, Zhou, Gong, Zhang, Yao, and
  Chen]{wang2022fca}
Donghua Wang, Tingsong Jiang, Jialiang Sun, Weien Zhou, Zhiqiang Gong, Xiaoya
  Zhang, Wen Yao, and Xiaoqian Chen.
\newblock Fca: Learning a 3d full-coverage vehicle camouflage for multi-view
  physical adversarial attack.
\newblock In \emph{Proceedings of the AAAI Conference on Artificial
  Intelligence}, pages 2414--2422, 2022.

\bibitem[Wang et~al.(2021)Wang, Liu, Yin, Liu, Tang, and Liu]{wang2021das}
Jiakai Wang, Aishan Liu, Zixin Yin, Shunchang Liu, Shiyu Tang, and Xianglong
  Liu.
\newblock Dual attention suppression attack: Generate adversarial camouflage in
  physical world.
\newblock In \emph{Proceedings of the IEEE/CVF Conference on Computer Vision
  and Pattern Recognition}, pages 8565--8574, 2021.

\bibitem[Xiao et~al.(2019)Xiao, Yang, Deng, Liu, and Li]{xiao2019meshadv}
Chaowei Xiao, Dawei Yang, Jia Deng, Mingyan Liu, and Bo Li.
\newblock {MeshAdv}: Adversarial meshes for visual recognition.
\newblock In \emph{Proceedings of the IEEE/CVF Conference on Computer Vision
  and Pattern Recognition}, pages 6898--6907, 2019.

\bibitem[Yamanaka et~al.(2020)Yamanaka, Matsumoto, Takahashi, and
  Fujii]{yamanaka2020depthpatch}
Koichiro Yamanaka, Ryutaroh Matsumoto, Keita Takahashi, and Toshiaki Fujii.
\newblock Adversarial patch attacks on monocular depth estimation networks.
\newblock \emph{IEEE Access}, 8:\penalty0 179094--179104, 2020.

\bibitem[Yang et~al.(2020)Yang, Chai, Anguelov, Zhou, Sun, Erhan, Rafferty, and
  Kretzschmar]{yang2020surfelgan}
Zhenpei Yang, Yuning Chai, Dragomir Anguelov, Yin Zhou, Pei Sun, Dumitru Erhan,
  Sean Rafferty, and Henrik Kretzschmar.
\newblock Surfelgan: Synthesizing realistic sensor data for autonomous driving.
\newblock In \emph{Proceedings of the IEEE/CVF Conference on Computer Vision
  and Pattern Recognition}, pages 11118--11127, 2020.

\bibitem[Yang et~al.(2022)Yang, Chen, Miao, Li, Zhu, and
  Zhang]{yang2022deepinteraction}
Zeyu Yang, Jiaqi Chen, Zhenwei Miao, Wei Li, Xiatian Zhu, and Li Zhang.
\newblock Deepinteraction: 3d object detection via modality interaction.
\newblock In \emph{Advances in Neural Information Processing Systems}, 2022.

\bibitem[Yang et~al.(2023)Yang, Chen, Wang, Manivasagam, Ma, Yang, and
  Urtasun]{yang2023unisim}
Ze Yang, Yun Chen, Jingkang Wang, Sivabalan Manivasagam, Wei-Chiu Ma,
  Anqi~Joyce Yang, and Raquel Urtasun.
\newblock {UniSim}: A neural closed-loop sensor simulator.
\newblock In \emph{Proceedings of the IEEE/CVF Conference on Computer Vision
  and Pattern Recognition}, pages 1389--1399, 2023.

\bibitem[Yin et~al.(2021)Yin, Zhou, and Krahenbuhl]{yin2021centerpoint}
Tianwei Yin, Xingyi Zhou, and Philipp Krahenbuhl.
\newblock Center-based 3d object detection and tracking.
\newblock In \emph{Proceedings of the IEEE/CVF Conference on Computer Vision
  and Pattern Recognition}, pages 11784--11793, 2021.

\bibitem[Zhang et~al.(2019)Zhang, Foroosh, David, and Gong]{zhang2019camou}
Yang Zhang, Hassan Foroosh, Philip David, and Boqing Gong.
\newblock {CAMOU}: Learning a vehicle camouflage for physical adversarial
  attack on object detectors in the wild.
\newblock In \emph{International Conference on Learning Representations}, 2019.

\bibitem[Zhang et~al.(2025)Zhang, Wang, Zhao, Zhao, Zhou, Dang, and
  Yin]{zhang2025dgaa}
Yixun Zhang, Lizhi Wang, Junjun Zhao, Wending Zhao, Feng Zhou, Yonghao Dang,
  and Jianqin Yin.
\newblock 3dgaa: Realistic and robust 3d gaussian-based adversarial attack for
  autonomous driving.
\newblock \emph{arXiv preprint arXiv:2507.09993}, 2025.

\bibitem[Zhang et~al.(2026)Zhang, Zhou, and Yin]{zhang2026bitaa}
Yixun Zhang, Feng Zhou, and Jianqin Yin.
\newblock {BiTAA}: A bi-task adversarial attack on 3d gaussian fields for
  cross-task perception.
\newblock \emph{IEEE Signal Processing Letters}, 33:\penalty0 1966--1970, 2026.

\bibitem[Zhao et~al.(2026)Zhao, Huai, Song, and Luo]{zhao2026cheatingstereo}
Kangqiao Zhao, Shuo Huai, Xurui Song, and Jun Luo.
\newblock Cheating stereo matching in full-scale: Physical adversarial attack
  against binocular depth estimation in autonomous driving.
\newblock In \emph{Proceedings of the AAAI Conference on Artificial
  Intelligence}, pages 13190--13198, 2026.

\bibitem[Zhu et~al.(2024)Zhu, Miao, Xue, Yu, Su, and Qiao]{zhu2024malicious}
Yi Zhu, Chenglin Miao, Hongfei Xue, Yunnan Yu, Lu Su, and Chunming Qiao.
\newblock Malicious attacks against multi-sensor fusion in autonomous driving.
\newblock In \emph{Proceedings of the 30th Annual International Conference on
  Mobile Computing and Networking}, pages 436--451. ACM, 2024.

\bibitem[Zhu et~al.(2023)Zhu, Zhang, Chen, Dong, Zhao, Ding, Zhong, and
  Zheng]{zhu2023bevrobustness}
Zijian Zhu, Yichi Zhang, Hai Chen, Yinpeng Dong, Shu Zhao, Wenbo Ding, Jiachen
  Zhong, and Shibao Zheng.
\newblock Understanding the robustness of 3d object detection with
  bird's-eye-view representations in autonomous driving.
\newblock In \emph{Proceedings of the IEEE/CVF Conference on Computer Vision
  and Pattern Recognition (CVPR)}, pages 21600--21610, 2023.

\end{thebibliography}
